\documentclass[10pt,twocolumn]{article} 
\usepackage{simpleConference}
\usepackage{amssymb}
\usepackage{latexsym}

\usepackage{url}
\usepackage{xcolor}
\definecolor{newcolor}{rgb}{.8,.349,.1}

\usepackage{amsmath,amssymb,amsfonts}
\usepackage{graphicx}
\usepackage{textcomp}

\usepackage[utf8]{inputenc} 
\usepackage[T1]{fontenc}    
\usepackage{hyperref}       
\usepackage{url}            
\usepackage{booktabs}       
\usepackage{amsfonts}       
\usepackage{nicefrac}       
\usepackage{microtype}      
\usepackage{lipsum}
\usepackage{framed,multirow}
\usepackage{amssymb}
\usepackage{latexsym}
\usepackage{makecell}
\usepackage{url}
\usepackage{xcolor}
\definecolor{newcolor}{rgb}{.8,.349,.1}
\usepackage{ragged2e}
\usepackage{color,soul}
\usepackage{hyperref}
\usepackage{graphicx}
\usepackage{caption}
\usepackage{subcaption}
\usepackage{float}
\usepackage{color}
\usepackage{algorithm}
\usepackage[noend]{algpseudocode}

\usepackage{array}
\newcolumntype{P}[1]{>{\centering\arraybackslash}p{#1}}

\usepackage{hyperref}
\usepackage{longtable}
\usepackage[switch,pagewise]{lineno} 

\begin{document}

\title{`CADSketchNet' - An Annotated Sketch dataset for 3D CAD Model Retrieval with Deep Neural Networks}

\author{\textbf{Bharadwaj Manda$^1$, Shubham Dhayarkar$^1$, Sai Mitheran$^2$,} \\ \textbf{V.K. Viekash$^2$, Ramanathan Muthuganapathy$^1$} \\
\small{$^1$ Indian Institute of Technology Madras $^2$ National Institute of Technology Tiruchirappalli}}

\maketitle
\thispagestyle{empty}

\begin{abstract}
Ongoing advancements in the fields of 3D modelling and digital archiving have led to an outburst in the amount of data stored digitally. Consequently, several retrieval systems have been developed depending on the type of data stored in these databases. However, unlike text data or images, performing a search for 3D models is non-trivial. Among 3D models, retrieving 3D Engineering/CAD models or mechanical components is even more challenging due to the presence of holes, volumetric features, presence of sharp edges etc., which make CAD a domain unto itself. The research work presented in this paper aims at developing a dataset suitable for building a retrieval system for 3D CAD models based on deep learning. 3D CAD models from the available CAD databases are collected, and a dataset of computer-generated sketch data, termed `CADSketchNet', has been prepared. Additionally, hand-drawn sketches of the components are also added to CADSketchNet. Using the sketch images from this dataset, the paper also aims at evaluating the performance of various retrieval system or a search engine for 3D CAD models that accepts a sketch image as the input query. Many experimental models are constructed and tested on CADSketchNet. These experiments, along with the model architecture, choice of similarity metrics are reported along with the search results. 
\end{abstract}


\section{Introduction}
\label{sec1}
The search or retrieval of Engineering (CAD) models is crucial for a task such as design reuse \cite{BAI20101069}. Designers spend a significant time searching for the right information and using a large percentage of existing design for a new product development \cite{Gunn82}. \cite{Ullman10} indicates that a large percentage (75\% or greater) of design reuses existing knowledge for the new product development. This calls for the search and classification of 3D Engineering models \cite{ShapeSearch}. With the wide applicability of 3D data and the increased capabilities of modelling, digital archiving, and visualization tools, the problem of searching or retrieving CAD models becomes a predominant one.

The research work presented in this paper aims at developing a well-annotated sketch dataset of 3D Engineering CAD models, that can aid in the development of deep learning-based search engines for 3D CAD models, using a sketch image as the input query. Using a sketch-based query for the search offers many advantages. 3D shapes, unlike text documents, are not easily retrieved using textual annotations (\cite{textquery1}) since it is difficult to characterize what human beings see and perceive using a mere text annotation. \cite{1314502, 4267943} show that content-based 3D shape retrieval methods (those that use the visual/shape properties of the 3D models) are shown to be more effective. It is also shown in \cite{1314502, 4267943} that using traditional search methods for multimedia data will not yield the desired results. \cite{944425} utilizes the idea of using the feature descriptors/vectors of the 3D model for the search query. Among the available query options, a sketch-based query is shown to be very intuitive and convenient for the user (\cite{6411819, 1268530, 2386318}), since it is easier for the user to learn and use such a system over using a 3D model itself as the query since it requires technical expertise and skill \cite{LI201457}. 

The Princeton Shape Benchmark (PSB) \cite{psb} was one of the earlier 3D shape databases. Consequently, large-scale datasets such as ShapeNet \cite{ShapeNet} came into being. Due to such data availability, many machine learning-based techniques, which require a good amount of data to train the models, have been developed (\cite{QiSMG17, PointNet++}). \cite{Eitz} provided the first benchmark dataset of sketches based on the 3D models in PSB. \textcolor{black}{As a result of this}, \cite{SHREC13} and \cite{SHREC14} have introduced large-scale benchmarks for sketch data for 3D shapes, including many approaches for sketch-based retrieval. 

\begin{figure}[t]
    \centering
    \begin{subfigure}[b]{0.22\textwidth}
        \centering
        \includegraphics[scale=0.1]{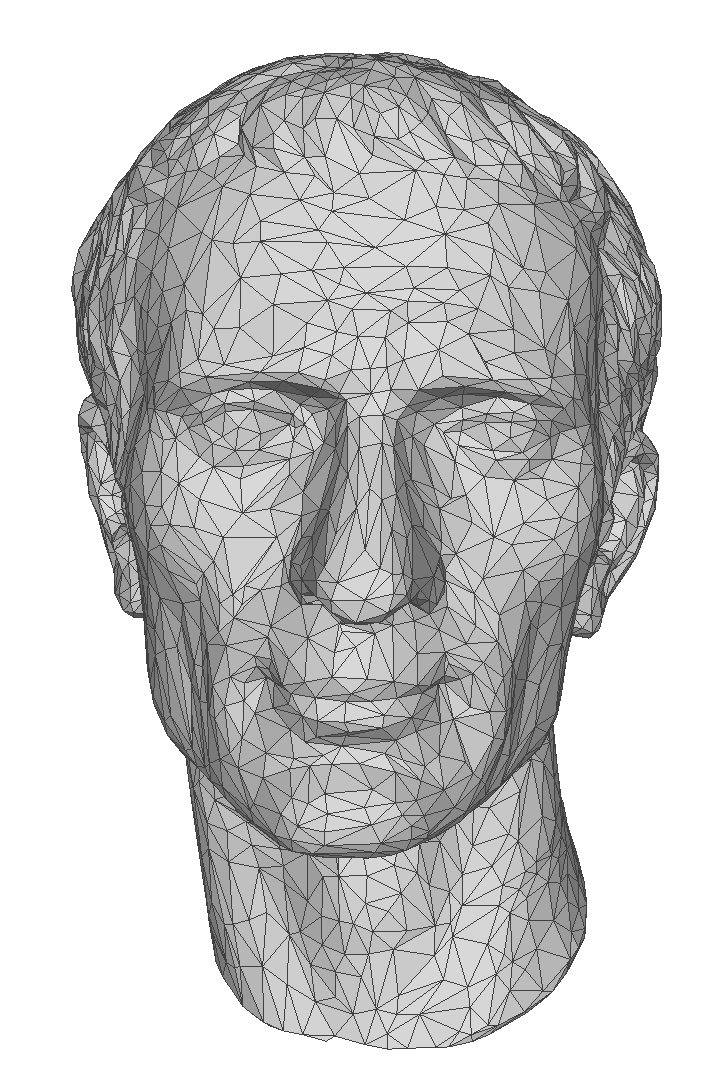}
        \caption{A 3D Graphical Model}
        \label{fig1b}
    \end{subfigure} \qquad
    \begin{subfigure}[b]{0.2\textwidth}
        \centering
        \includegraphics[height=3.5cm,width=4cm]{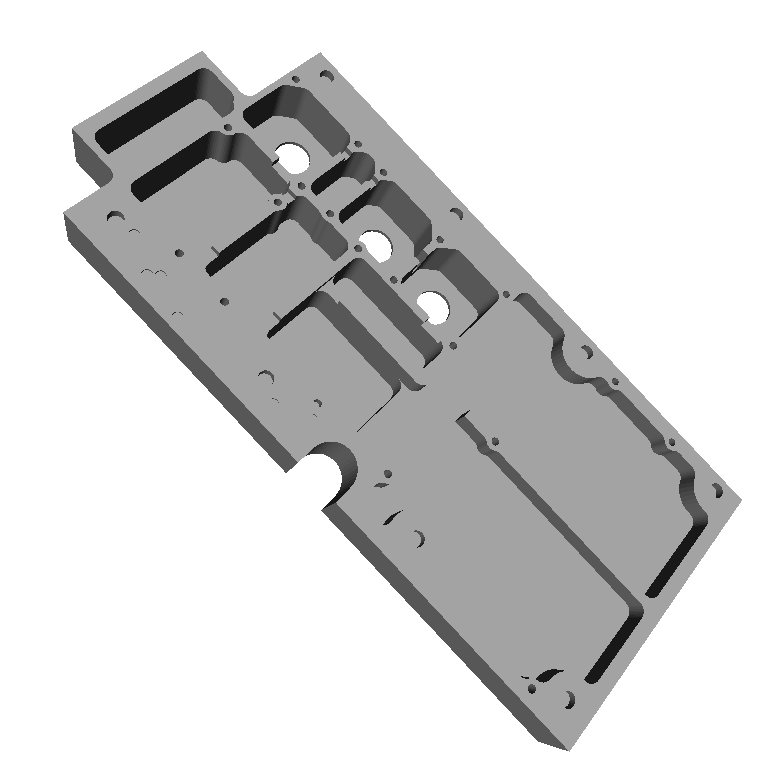}
        \caption{A 3D CAD Model}
        \label{fig1a}
    \end{subfigure}
    \caption{Distinction between a Graphical Model and a CAD Model}
\end{figure}

However, these datasets and methodologies only aim at generic 3D shape data or graphical models and do not contain Engineering CAD models. The presence of features such as holes (genus $>$ 0), blind holes (genus $=$ 0) and other machining features in an Engineering/CAD model (see Fig. \ref{fig1a}) calls for special treatment as opposed to a 3D graphical model where such features and holes are usually absent (see Fig. \ref{fig1b}). Also, sharp edges are usually found in a CAD Model \textcolor{black}{unlike} a graphical model that has smooth curvature throughout. 

In the case of Engineering models, the people involved are required to have rich domain knowledge and experience. CAD models are typically obtained via the design process, unlike 3D shape data which are acquired via 3D scans. Since most design data are proprietary, they are not available to the public domain (e.g. \cite{Qin2014}). Moreover, among the few datasets that are available for usage, they either lack proper annotations \cite{ABC} or contain too few models (\cite{IpRegli, Wu, NDR}), \textcolor{black}{which are} not very conducive for employing deep learning methods. Although the work presented in \cite{Qin2017} performs a sketch-based retrieval of 3D CAD models, the dataset is proprietary. Engineering shape benchmark (ESB) \cite{ESB} is a prominent dataset for CAD models having 801 models. More recently, datasets such as mechanical components benchmark (MCB) \cite{mcb} and CADNET \cite{CADNET} have been proposed. Nonetheless, the data is in the form of 3D models, and no sketch information is present. As far as the sketch-data for Engineering models is concerned, there are no datasets available to the best of our knowledge. 

Our motivation for addressing the problem of retrieval of Engineering / CAD models, therefore, comes from the following:
\begin{enumerate}
\item Most CAD datasets are typically contain only a few models and hence cannot be utilized by the latest technological advances such as deep neural networks. 
\item Datasets having larger number of CAD models are either proprietary  (not publicly available) \cite{Qin2014} or lack classification information \cite{ABC}.
\item There is no existing database for sketch-data of 3D CAD models since obtaining hand-drawn sketches for CAD models is difficult.
\item The recent advances in deep learning have not been made use of, to the best of our knowledge. 
\end{enumerate}

The key contributions of the paper are:
\begin{enumerate}
\item Using the available 3D CAD models from \cite{ESB} and \cite{mcb}, a large-scale dataset of computer-generated sketch data called `CADSketchNet' is created.
\item Additional hand-drawn sketches of CAD models are also incorporated into the dataset using the models available from \cite{ESB}.
\item To benchmark the developed dataset, we analyzed the performance of various learning-based approaches for the sketch-based retrieval of 3D CAD models.
\item The performance of the various experimental retrieval systems on CADSketchNet are compared, and the search results are reported. 
\end{enumerate}

{\color{black}This paper makes useful contribution to the research community involving `mechanical components' and allow researchers to develop new algorithms for the same.} The paper is organized as follows: Section \ref{sec2} discusses the related works corresponding to 3D CAD models, in addition to the literature on images and generic 3D Shapes. The dataset preparation is explained in Section \ref{sec4} including the challenges involved in creating hand-drawn sketch data for 3D CAD models, the need for computer-generated sketches and the process of generating such sketches. The experiments done in order to benchmark the developed `CADSketchNet' dataset are detailed in Section \ref{sec:exp}. Section \ref{sec6} provides the Implementation Details. The results, limitations and possible future work are elaborated in Section \ref{sec:res}, followed by a Conclusion (Section \ref{sec:concl}). The dataset is available at \url{https://bharadwaj-manda.github.io/CADSketchNet/}

\section{Related Works}
\label{sec2}
Many works in recent times have focused on 3D graphical models and images. We focus more on the approaches that have been proposed for the search and classification of 3D mechanical components, which are very few. However, a few of the recent approaches used in other domains are also mentioned for the sake of completeness.

\subsection{Images}
Deep learning became ubiquitous for image-related tasks since the application of Convolutional Neural Networks (CNN) to the ImageNet Challenge \cite{Russakovsky2015}. For the sketch-based retrieval of image data, sketch datasets such as \cite{sketchy} and \cite{TUBerlin} have been introduced. The work in \cite{Radenovic_2018_ECCV} extracts edge maps of the matching and non-matching images and uses these maps for training. \cite{8953251}, \cite{Yelamarthi2018AZF} popularize the zero-shot learning framework. \cite{Bhunia2020SketchLF} attempts an on-the-fly sketch-based image retrieval using reinforcement learning methodology. 

\subsection{3D models of common objects}
\cite{ShapeNet} introduced the ShapeNet dataset and also performed classification and retrieval tasks on it. Many learning-based approaches have been proposed for the tasks of classification and retrieval using the ModelNet dataset. The leader-board can be found at \cite{ModelNet}, with the popular methods being \cite{MVCNN, 3dor.20171045, 7353481}. However, none of them uses a sketch query. A few tracks of the Shape Retrieval Contest (SHREC) involved a sketch-based retrieval challenge for 3D shapes. SHREC'12 \cite{SHREC12} involved a sketch-based retrieval task and introduced a benchmark dataset. Building upon the dataset by \cite{Eitz}, SHREC'13 \cite{SHREC13} included a Large Scale Sketch-Based 3D Shape Retrieval track with a larger dataset. \cite{LI201457} compares different methods for sketch-based 3D shape retrieval. SHREC'14 \cite{SHREC14} expanded upon this further to an Extended Large Scale Sketch-Based 3D Shape Retrieval track. SHREC'17 \cite{SHREC17} involved a sketch-based retrieval task of 3D indoor scenes.

\subsection{3D CAD models of Engineering shapes}
The SHREC track \cite{SHREC08} presents a retrieval challenge using the Engineering Shape Benchmark (ESB) dataset \cite{ESB}. \cite{1779120} uses the idea from content-based image retrieval to the domain of 3D CAD models. \cite{PU2006249} performs a visual similarity-based retrieval using 2D engineering drawings. This method converts 2D drawings to a shape histogram and then applies the idea of spherical harmonics to obtain a rotation-invariant shape descriptor. Minkowski distance was used to measure the similarity between feature representations. 

\cite{2007.04.005} uses the data from ShapeLab \cite{ShapeLab} and ESB to develop a sketch-based 3D part retrieval system. This paper uses the idea of classifier combinations to aid in the retrieval process. Engineering models are classified functionally and not visually, as opposed to the image or 3D graphical data. Taking this into account, the extracted shape descriptors (Zernike Moments and Fourier Transforms) are sent to a Support Vector Machine (SVM) classifier. A weighted combination of the classifier outputs is then used to estimate the class or category of the input query, which is then compared against the classes of the database. This is one of the earlier works that use learning-based methods for building a sketch-based retrieval system for CAD models. However, the sketch-data itself is not available.

More recently, a sketch-based semantic retrieval of 3D CAD models is presented by \cite{Qin2017}. The CAD models and their parametric features (used in 3D modelling software) are taken. The pre-processed sketch query is first vectorized and then passed onto topology-based rules. An integrated similarity measurement strategy is used to compute the similarity between the query sketch and CAD models' database. This research work uses a dataset of 2148 CAD models and six corresponding views of each model. This dataset is proprietary and is not available. Also, the method presented here uses the classical rule-based approach over the latest advances in learning-based approaches. 

{\color{black} 
\subsection{Other related works}
\cite{bonnici2019sketch} presents a detailed study on the state-of-the-art methods and the future of sketch-based modelling and interaction. However, the discussion mainly focuses on sketch interpretation and on the development and usage of interactive sketches. There is very limited discussion on 3D engineering shape data.

\cite{OpenSketch19} introduces the OpenSketch Dataset which contains annotated sketches corresponding to various product designs. A detailed study is done in order to understand the stroke time and pressure. A taxonomy of lines is also provided and the strokes are labelled. However, the dataset contains only 107 sketches across 12 categories, which is not sufficient for developing learning based models. 

The SPARE3D dataset \cite{SPARE3D} aims at understanding the spatial reasoning behind line drawings using deep neural networks. While the dataset uses 10,369 3D CAD models, it only contains line drawings of 3D objects from 8 different isometric views and does not contain any hand-drawn sketches. ProSketch3D \cite{ProSketch} consists 1500 sketches of 3D models across 500 object categories taken from ShapeNet. All sketches corresponding to generic 3D shapes and not 3D mechanical components.}

\section{Dataset Creation}
\label{sec4}
As discussed in Section \ref{sec2}, a database of 3D models and their corresponding sketches are not available for the domain of CAD models. \cite{SketchGraphs} introduces a sketch-dataset for CAD models. However, it is not useful for a traditional search problem since the sketches are based on the design workflow, i.e. based on `how' a model is designed, rather than the model shape and geometry. \textcolor{black}{\cite{Qin2017} uses only a proprietary dataset.} Therefore, we attempt to build a new sketch-dataset, \textcolor{black}{termed `CADSketchNet',} using the 3D models from existing CAD databases. 

\subsection{Challenges in creating a dataset of hand-drawn sketches}
\label{sec:challange}
Creating a hand-drawn sketch for a 3D mechanical component is much more challenging than sketching a generic 3D shape. This is because,

\begin{itemize}
    \item It is difficult to capture the detailed information present in a CAD model, such as the presence of holes and volumetric features in a single sketch.
    \item Multiple viewing directions can be chosen to draw the sketch.
    \item The sketches need to be drawn by users with domain knowledge and experience. Gathering a set of users to contribute to building such a dataset is both time-consuming and expensive.
    \item Once the hand-drawn sketches are obtained, they need to be verified and validated for correctness and closeness to the input CAD model.
    \item Different users have different drawing styles, and consistency needs to be maintained across the hand-drawn sketches.
\end{itemize}

Due to these reasons, attempting to create a dataset of hand-drawn sketches for a large number of 3D CAD models is a tedious task.

\subsection{Hand-drawn sketch data generation}
\label{sec:hand-drawn}
The ESB dataset \cite{ESB} is a publicly available CAD database that is also well annotated. In the ESB, there are 801 3D CAD models across 42 classes (excluding the models in the `Misc' category). {\color{black} Since, the ESB is a reasonably sized dataset, we attempt to obtain hand-drawn sketches for all 801 3D CAD models of the ESB.

\subsubsection{Gathering users for obtaining hand-drawn sketches}}
\textcolor{black}{Around 50 users with experience of CAD and engineering drawing were selected in order to provide a hand-drawn sketch for each 3D CAD model in the ESB. The users are mostly engineering students pursuing undergraduate and postgraduate courses in design, with a few industry professionals. Because the goal is to create a dataset that can be used to develop learning-based solutions, users were asked to draw the sketches in their natural style so that the dataset might capture more variability. Furthermore, users were encouraged not to be overly precise and correct, because the learning algorithms that could use this dataset can capture input variations while staying robust to noise.}

\subsubsection{Obtaining the hand-drawn sketches}
Each user was then shown the 3D object (digital) and is asked to draw a digital sketch on a hand-held tablet device. Since a 3D object can be viewed from many directions, the best viewing direction for the sketch (i.e. that which covers the entire geometry of the object) is determined by the user, based upon the domain knowledge and experience. {\color{black} The cases of potential ambiguity with respect to the viewing direction are resolved by a majority vote among the users.} 

\textcolor{black}{As we only attempt to draw a single sketch for each 3D CAD model in the ESB dataset, the number of sketches obtained and the category information acquired are the same as that of the ESB. The reader can refer to the paper on ESB \cite{ESB} for more information on the category information and the number of models in each class.}

\subsubsection{Processing the hand-drawn sketches}
Initial cleaning of these hand-drawn sketches was done by sketch pre-processing using the software in Autodesk Sketchbook. Consequently, the sketches were then validated for their correctness and closeness (resemblance to the 3D object) by domain experts from academia and the industry. These images and the class labels (the same label as the 3D model from ESB) are stored as a database. Hence, the number of sketches is only of the order of few hundreds. Nevertheless, this dataset is stored and will be made available since it is challenging to obtain a real dataset of hand-drawn sketches. This dataset could also be used by algorithms that do not need large-scale training data. 

\begin{figure}[t]
    \centering
    \includegraphics[scale=0.17]{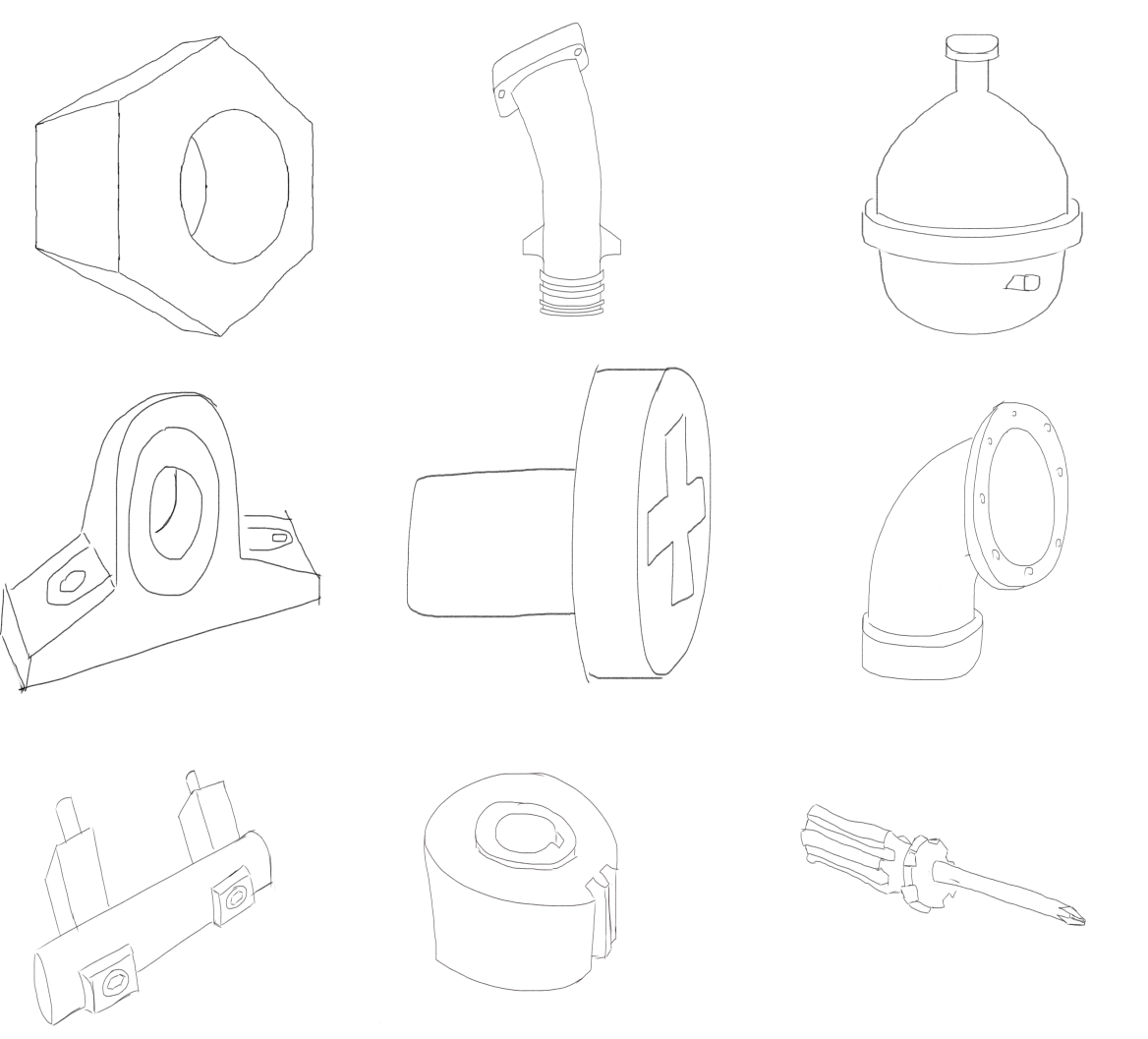}
	\caption{Sample Hand-drawn sketches from the developed `CADSketchNet'.}
    \label{fig:hand-drawn}
\end{figure}

{\color{black} \subsubsection{Analyzing the hand-drawn sketches}
During the course of obtaining user drawn sketches, it was observed that simple object classes like Clips, Bolt-Like Parts, Nuts and Discs were very easy to draw, \textcolor{black}{and took very little time to obtain. Users encountered significant difficulty when they were required to sketch object categories with complex features, such as 90-degree elbows, Motor Bodies, Rectangular Housing, and so on.} For most other object categories, the users were able to draw the sketches with a manageable level of difficulty.

\textcolor{black}{There is also a need to assess the quality of the sketches in terms of the correctness of each user's choice of viewing direction. The concept of Light Field Descriptor (LFD) is proposed by \cite{LFD}, in which a 3D model is placed inside a regular dodecahedron and images are obtained by using each vertex as a viewing direction. Using this idea, 20 view images are obtained for each 3D object.}

The obtained sketch is compared with each of the 20 view images. The view image with the highest obtained similarity score is noted, and the viewing direction is cross-checked with the user's choice of viewing direction. In majority of the cases, the gathered users were able to \textcolor{black}{accurately determine the ideal viewing direction for a model. In the few circumstances where this was not the case, the user was requested to redraw the sketch. The quality of the produced hand-drawn sketch dataset is thus ensured. Figure  \ref{fig:hand-drawn} shows a few sample sketches.}}

\subsection{Creating Computer-generated Sketch Data}
\label{sec:synt}
\textcolor{black}{The Mechanical Components Benchmark (MCB) \cite{mcb}, contains 58,696 3D CAD models across 68 classes. Hence, for our study,} we utilize the 3D CAD models from the MCB to prepare a dataset of sketch images. Deep learning methodologies are data-driven, and it calls for a large amount of data. Since it is tough to obtain hand-drawn sketches for such a large-sized dataset of complex 3D mechanical components, we attempt to create computer-generated sketch images corresponding to each CAD model in the MCB.

\begin{figure*}
	\centering
	\includegraphics[width=1\textwidth,height = 4.5cm]{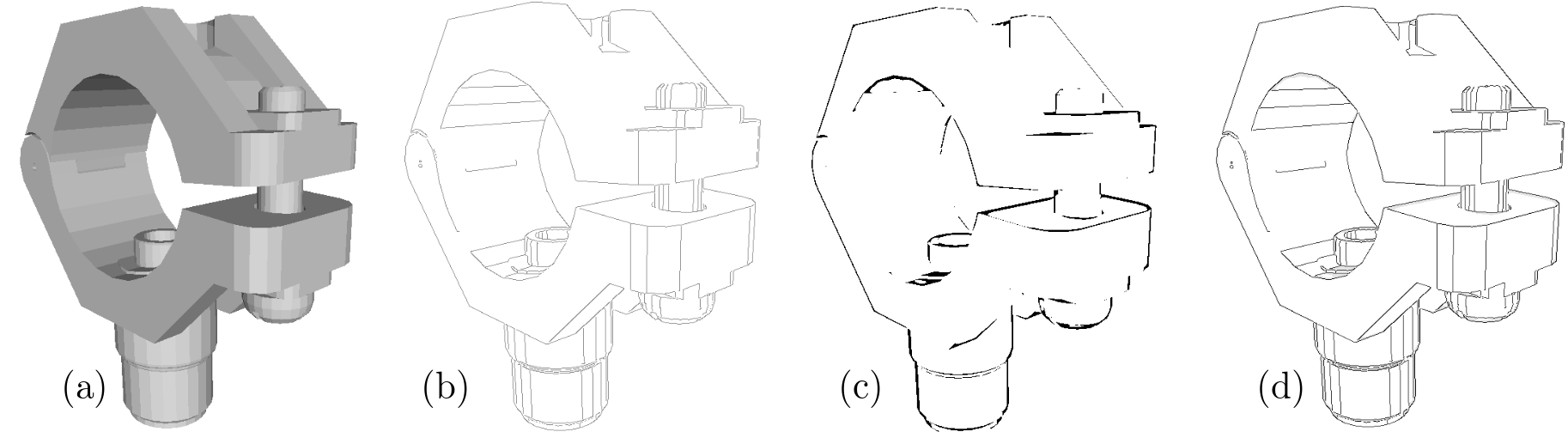}
	\caption{(a) 2D image obtained from Algorithm \ref{alg1} (b) Result of Canny edge detection (c) Result of Gaussian blurring (d) Result of weighted scheme of (b)\&(c)}
	\label{fig:edge}
\end{figure*}

\subsubsection{Converting 3D model to a 2D image}
\label{sec:synt1}
A representative 2D image of every 3D model in the MCB needs to be obtained as a first step. A Python script is used to save the image of the 3D model from a particular viewing direction. \textcolor{black}{Applying the idea of LFD,} we obtain 20 images for every 3D object. Now it is needed to identify one representative image for each object. For this task, a group of 70 volunteers with knowledge of CAD were identified. The 3D CAD objects were split into batches. For each batch, the above procedure was applied, and the one among the resultant output images was chosen as the representative image for the 3D object. Repeating this process for every batch, one image corresponding to every 3D CAD model in the MCB was generated. The overall procedure is summarized in Algorithm \ref{alg1}.

\begin{algorithm} 
\caption{Method to obtain a 2D image for a 3D model}
\hspace*{\algorithmicindent} \textbf{Input} Database of 3D CAD models
\begin{algorithmic}[1]
\Procedure{3dobj\_to\_img}{}       
	\State Split the 3D models in the database into $n$ batches
    \State $i = 1$
	\While{$i <= n$}  
        
	\For{\texttt{every model in the batch}}
        \State \texttt{Apply LFD to obtain 20 images}
        \State \texttt{Users identify 1 among 20 images}
      \EndFor    
    \EndWhile   
\EndProcedure
\end{algorithmic}
\hspace*{\algorithmicindent} \textbf{Output} 2D representative image for every 3D CAD model	
\label{alg1}
\end{algorithm}

\subsubsection{Creating computer-generated sketch from the image}
\label{sec:synt2}
{\color{black} There is ample literature related to generating computer sketches. Edge detection of images is a fundamental approach to obtaining the object boundaries. Hence, we first begin our experiments to generate computer sketches for the 3D CAD models in the MCB dataset with the popular edge detection methods.

We first experiment with the Canny edge detection method \cite{Canny}, which essentially tries to identify the object boundaries present in the image using first derivative methods to identify local image features. The method is applied to the 2D images of the CAD models obtained from Algorithm \ref{alg1}. In order to enhance the obtained edges, we combine the Canny edge detection process with a second approach that uses the idea of image blurring, coupled with boundary shading. By using Gaussian blurring, the internal object features are suppressed. Thus, a weighted combination of the Canny edge detection and the idea of Gaussian blurring is proposed for generating the computer sketches.}

\begin{algorithm} 
\caption{Method to generate computer sketch from image}
\hspace*{\algorithmicindent} \textbf{Input} Images of 3D CAD models from Algorithm \ref{alg1}
\begin{algorithmic}[1]
\Procedure{img\_to\_sketch}{}      
	\State $I \leftarrow$ Read input image in RGB color space
    \State $G \leftarrow$ Convert image from RGB colorspace to Grayscale 
    \State $IG \leftarrow$ Invert color values of all pixels in Grayscale
    \State $B \leftarrow$ Convolve non-uniform GF : kernel size $k$ \& SD $\sigma$ 
    \State $IB \leftarrow$ Invert the blurred image
    \State Element-wise division of $G$ \& $IB$ with $scale=256.0$
    \State $O1 \leftarrow$ Binary threshold the obtained image
    \State $O2 \leftarrow$ Canny Edge Detection of $G$
    \State $S \leftarrow$ Weighted average over O1 and O2 
\EndProcedure
\end{algorithmic}
\hspace*{\algorithmicindent} \textbf{Output} Computer-generated sketches of 3D CAD models
\label{alg2}
\end{algorithm}

The reason for using these two techniques is as follows. Since the original images are obtained from CAD mesh models, they contain several mesh lines. To generate sketches that can capture the overall shape and geometry of the object, we need to use methods capable of extracting the significant edges without paying too much attention to minute details of the object. Usage of the canny filter detects prominent edges effectively by thresholding, even in a noisy environment. The minute noisy details are easily removed due to the presence of a Gaussian Filter (GF), and the required signal (prominent pixels) can be enhanced using the Canny edge detector that uses non-maximum suppression against the noise, resulting in a well-defined output. Additionally, to enhance the output, we operate only in gray-scale rather than RGB. The procedure is summarized in Algorithm \ref{alg2}.

The weight assigned for the Gaussian blurring is significantly less in order to avoid too much suppression of minute details and thus leading to loss of vital information. Hence, the output of the weighted scheme closely resembles the output of the Canny edge detection scheme. The outputs of both these methods, along with the output of the weighted scheme, are shown in Figure \ref{fig:edge} for a sample input.

{\color{black} In addition to the weighted Canny edge detection method mentioned above, other popular edge detectors such as the Scharr, Prewitt, Sobel and Robert Cross operators are also experimented with. A similar weighted scheme is applied to each of these methods. The other state-of-the-art sketch generation methods such as NeuralContours \cite{neural_contours}, PhotoSketch \cite{photo}, and Context-aware tracing strategy (CATS) \cite{CATS}, are also experimented.}

\renewcommand{\arraystretch}{1.2}
\begin{table*}[t]
\centering
\resizebox{\textwidth}{!}{%
\begin{tabular}{|P{4cm}|c|c|c|c|c|c|P{3cm}|}
\hline
\textbf{Sketch-generation method}& \textbf{PSNR ↑} & \textbf{MS-SSIM ↑} & \textbf{IE ↑} & \textbf{VIF ↑} & \textbf{MSE ↓} & \textbf{UQI ↑} & \textbf{Conversion time (Per image in sec) ↓} \\ \hline
plain-canny & 18.0834 & 0.5718 & 1.5248 & 0.0034 & 1010.9600 & 0.9874 & \textbf{0.0021} \\ \hline
weighted-scharr & 21.3913 & 0.6327 & 1.3649 & 0.0031 & 472.0136 & 0.9948 & 0.0581 \\ \hline
weighted-prewitt & 21.7143 & 0.6412 & 1.3904 & 0.0031 & 438.1824 & 0.9952 & 0.0547 \\ \hline
weighted-roberts & 20.4433 & 0.6169 & 1.3498 & 0.0031 & 587.1513 & 0.9935 & 0.0517 \\ \hline
weighted-sobel & 21.6066 & 0.6388 & 1.3555 & 0.0031 & 449.1815 & 0.9951 & 0.0498 \\ \hline
neural-contours \cite{neural_contours} & \textbf{25.4318} & \textbf{0.9319} & 1.3425 & \textbf{0.5292} & \textbf{186.1659} & \textbf{0.9977} & 828.50 \\ \hline
photosketch \cite{photo} & 12.5434 & 0.4978 & \textbf{3.7558} & 0.0055 & 3620.2508 & 0.9367 & 9.0180 \\ \hline
CATS \cite{CATS}  & 16.2040 & 0.5428 & 1.5796 & 0.0035 & 1558.3710 & 0.9788 & 1.7810 \\ \hline
\textbf{weighted-canny (ours)} & 24.9429 & 0.8208 & 1.6737 & 0.0034 & 209.4152  & \textbf{0.9977} & 0.0190 \\ \hline
\end{tabular}%
}
\caption{Similarity results obtained by using various approaches for comparing the hand-drawn and the computer-generated sketches for various sketch-generation methods \textcolor{black}{on the 801 CAD models of the ESB dataset}. ↑ indicates that greater value for the metric indicates higher similarity, while ↓ indicates the opposite. \textcolor{black}{The plain-canny results reported here are with non-maximal suppression.} The similarity measures used are PSNR - Peak signal-to-noise ratio; MS-SSIM - Multi Scale Structural Similarity Index \cite{MSSSIM}; IE - Information Entropy; VIF - Visual Information Fidelity \cite{VIF}; MSE - Mean Squared Error; UQI - Universal image Quality Index \cite{UQI};}
\label{tab:simi}
\end{table*}
\renewcommand{\arraystretch}{1.00}

\subsubsection{Comparing hand-drawn and computer-generated sketches}
\label{sec:synt3}
Creating hand-drawn sketches for a 3D CAD model is extremely difficult, as established in Section \ref{sec:challange}, and since such sketch data is not available for the MCB, we cannot directly compare the computer-generated and hand-drawn sketches. However, for the ESB dataset, we have obtained hand-drawn sketches (see Section \ref{sec:hand-drawn}). {\color{black} Therefore, computer sketches are generated for the 801 models in ESB, and these are compared with the corresponding hand-drawn sketches.} 

{\color{black} Many sketch generation methods were experimented in Section \ref{sec:synt2}. To find out which among these methods result in sketches that most closely resemble the hand-drawn ones, we attempt to compare the computer generated sketch and the hand-drawn sketch of each 3D CAD model in the ESB. Various state-of-the-art similarity metrics are used, and the average similarity score across all models is obtained. A detailed comparison is reported in Table \ref{tab:simi}. From the Table, it is clear that the proposed weighted canny approach performs much better than the plain Canny edge detection. \textcolor{black}{The MSE value with and without the non-maximal suppression (NMS) stage of the plain canny method are 1010.96 \& 1577.43 respectively, while that of weighted canny is 209.41. The values indicate that canny without NMS performs poorly, and the weighted canny approach performs the best.}

\begin{figure*}[t]
	\centering
	\includegraphics[width=1\textwidth, height = 3.5cm]{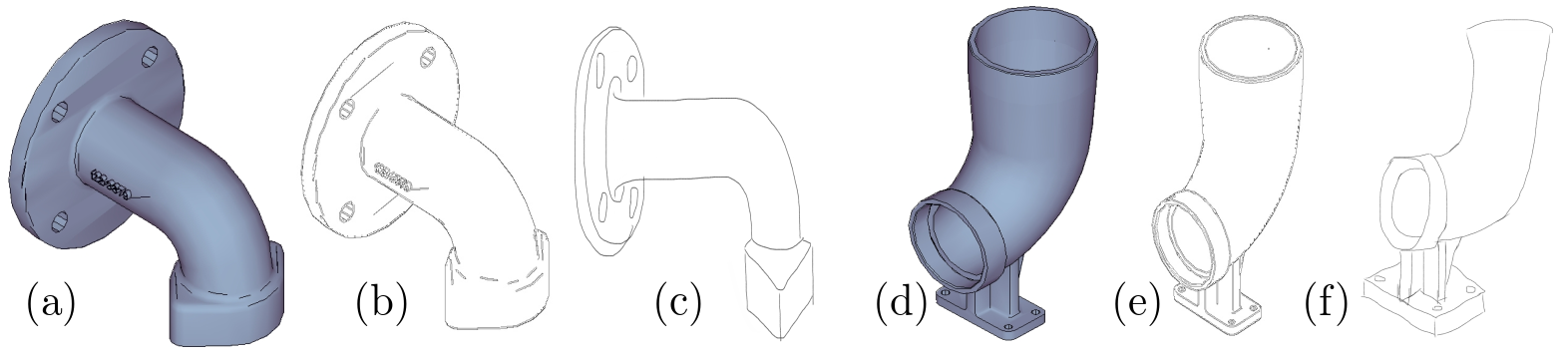}
	\caption{(a), (d) - Sample Images extracted from two random CAD models in ESB; (b), (e) Computer generated sketch data; (c),(f) - hand-drawn sketch data; Although the computer-generated sketch has a lot of detail and resembles the input closely, the hand-drawn sketch database provides realistic query images that aid in training robust retrieval systems}
    \label{fig:hand-synt}
\end{figure*}

It can also be seen that for most similarity metrics, NeuralContours \cite{neural_contours} generates a sketch closest to the hand-drawn sketch. However, the time taken to generate one sketch through the other methods is much higher than the proposed sketch generation method in Section \ref{sec:synt2}. The neural contours method takes around 800 seconds to generate a single sketch image on an NVIDIA 1080Ti GPU, owing to the complex neural network pipeline. This is not suitable for generating sketches for a large number of 3D models. On the other hand, while the performance of the proposed weighted-canny method is close enough to the Neural Contours method in most of these cases, the time taken by to generate one sketch from a 3D model is just one second (including converting a 3D model to an image \textcolor{black}{followed by generating a sketch}), on the same hardware setup. This aids very much in generating sketches for a large dataset of 3D models such as the MCB. Hence, the proposed method of weighted canny edge detection is chosen to efficiently generate all the computer sketches of the MCB dataset.}

\begin{figure*}[t]
	\centering
	\includegraphics[width=.85\textwidth, height = 7.5cm]{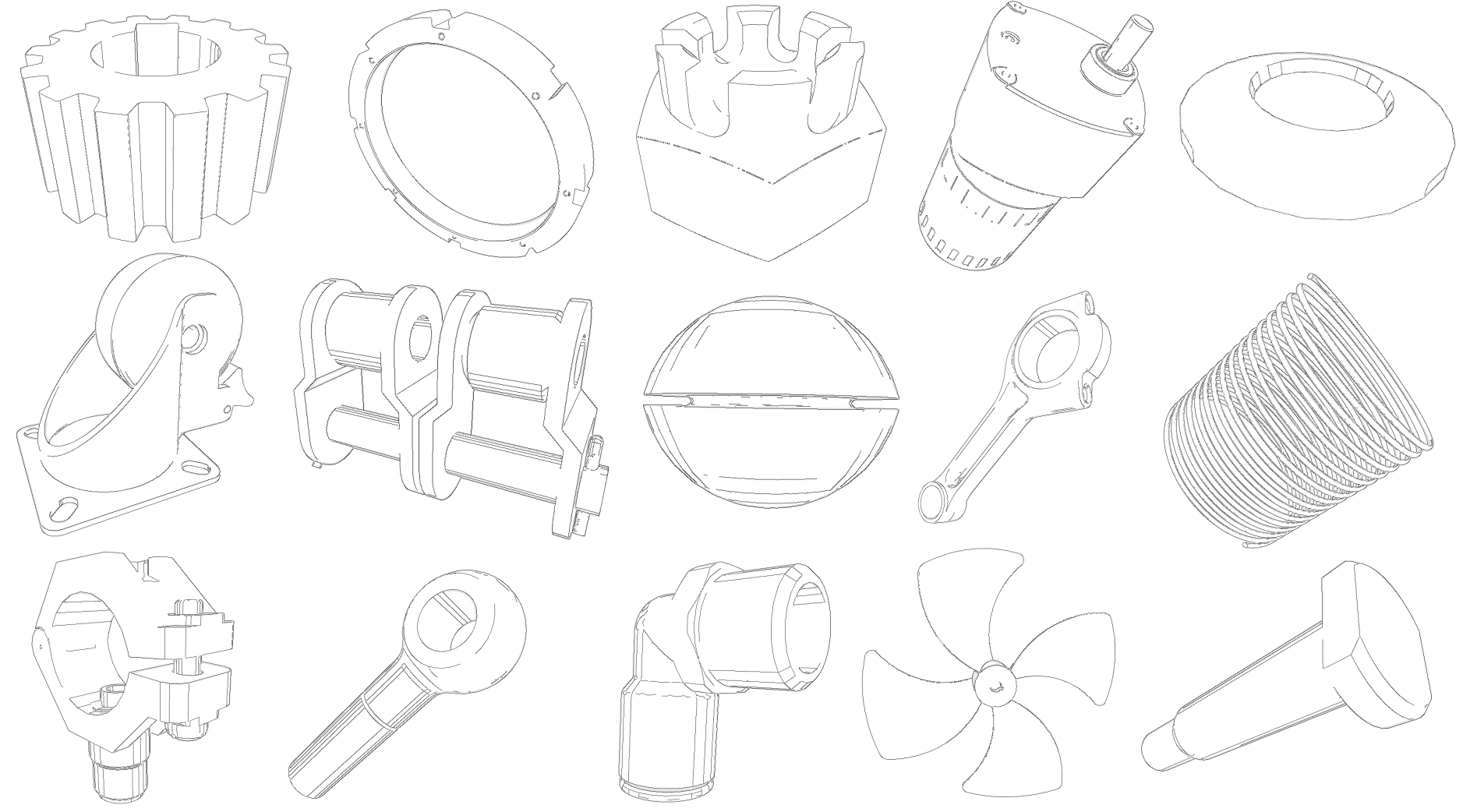}
	\caption{Sample images from the developed `CADSketchNet' Dataset-A: Computer-generated sketches.}
    \label{fig:comp-gen}
\end{figure*}

It is important to note that the goal of creating this dataset is to aid in the development of deep learning-based CAD model search engines. Since the end-users of the search engine are humans, and sketches drawn by an average human being are bound to have errors, the dataset needs to contain sketches that are not perfect. Only then will the learning-based methodologies that make use of this dataset become robust to input noise and errors. Figure \ref{fig:hand-synt} shows, for two sample cases, the image of the 3D CAD model, the hand-drawn sketch and the computer-generated sketch. It can be observed that while the computer-generated sketch \textcolor{black}{contains a lot of detail} and bears a close resemblance to the input, the sketch does not look realistic. On the other hand, the hand-drawn sketches provide a realistic database of query sketches that can be used to train a robust search engine. Clearly, a hand-drawn sketch dataset is more important and valuable as compared to computer-generated sketch data. Nevertheless, in the absence of a standard large-scale benchmark dataset of hand-drawn sketches, the computer-generated sketch data generated for the MCB dataset using Algorithm \ref{alg2} is the best option available.

\begin{figure*}
	\centering
	\includegraphics[scale=0.34]{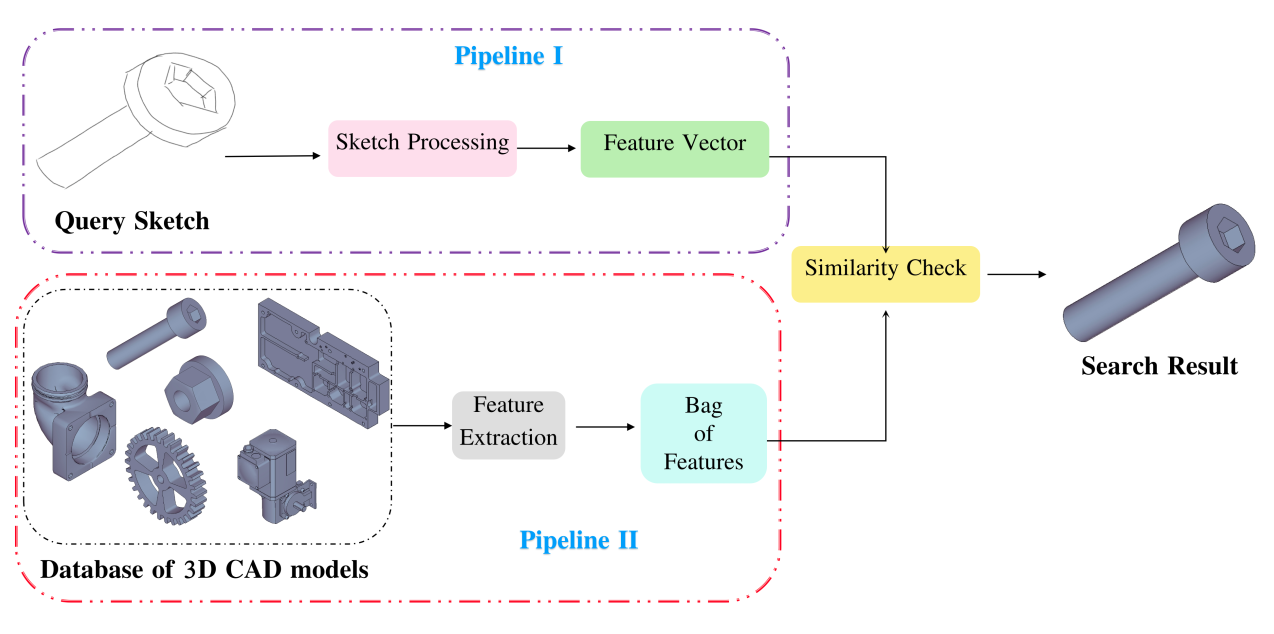}
	\caption{A generic pipeline for a sketch-based 3D CAD model search engine. Input query - a sketch image drawn by the user. Pipeline I - input sketch pre-processing followed by feature extraction. Pipeline II - extracts features of all the CAD models in the database and stores them as a bag of features. The output feature vector from I is compared against the bag of features from II. Using similarity metrics, the object(s) that is(are) most similar to the input query is(are) retrieved.}
	\label{fig2}
\end{figure*}

\subsubsection{Analysis of the proposed sketch-generation method}
{\color{black} To further understand the complexity involved in generating the sketch data, the time taken by the proposed technique is computed for each class of the MCB. Since the MCB dataset is not class-balanced, i.e. the data is unevenly distributed across classes, we compare the average time for generating a single sketch of a particular class.} It is observed that complex object categories such as Helical Geared Motors, Castor, Turbine etc. take a significantly higher time compared to the other categories. Simple object classes such as Convex Washer, Cylindrical Pin, Setscrew, Washer Bolt etc., take negligible time.

As discussed earlier with respect to the hand-drawn sketches, we only attempt to generate a single sketch corresponding to every 3D CAD model in the MCB dataset and do not change anything else. Hence, the number of sketches obtained are the same as the number of CAD models in MCB, i.e. 58696 sketches across the 68 classes of the MCB. For detailed information regarding the categories and the number of models in each class, the reader is directed to the MCB paper \cite{mcb}.

\subsection{Summary and `CADSketchNet' Details}
The dataset `CADSketchNet' contains two subsets. 
\begin{itemize}
	\item Dataset-A contains (1) one representative image for each 3D CAD model in the MCB dataset (obtained using Algorithm \ref{alg1}) (2) one computer-generated sketch for each representative image in the MCB dataset (obtained using Algorithm \ref{alg2}). This results in 58,696 computer-generated sketches across 68 categories. {\color{black} Some sample sketches generated by the proposed method are shown in Figure \ref{fig:comp-gen}.}

	\item Dataset-B contains 801 hand-drawn sketches, one for each 3D CAD model in the ESB dataset across 42 categories (as described in Section \ref{sec:hand-drawn}). {\color{black} Since the 3D CAD model data is obtained from the ESB, the same category information applies to the Dataset-B as well. Some sample hand-drawn sketches are shown in Figure \ref{fig:hand-drawn}.}
\end{itemize}

Dataset-A has all the images split into an 80-20 ratio for training and testing, respectively. This split is as per the MCB \cite{mcb}. Dataset-B contains no train-test split since the size of the dataset is not as large as Dataset-A. Nonetheless, users who intend to use this data can customize the train-test ratio as needed.

\section{Experiments}
\label{sec:exp}
In this section, we analyze the behaviour of a few learning algorithms for 3D CAD model retrieval on the Dataset-A and Dataset-B of the CADSketchNet.

\color{black}
\subsection{Experiments on Dataset-A}
\label{data-A}
Methods proposed in literature mainly use point cloud representations (\cite{QiSMG17,PointNet++}), and Voxel-grid representations (\cite{VoxNet}). {\color{black} These network architectures take in point cloud inputs or graph inputs etc., and not images. Since the sketch data is available in the form of images, it is not possible to experiment with these architectures. However, a few papers use view-based representations (\cite{MVCNN}, \cite{GVCNN}).} Since, we are dealing mainly with the image representations of 3D CAD models, only the view-based methods can be experimented on the Dataset-A of `CADSketchNet'. 

The performances of four view-based learning architectures are analysed: MVCNN \cite{MVCNN}, GVCNN \cite{GVCNN}, RotationNet \cite{kanezaki2021_rotationnet}, and MVCNN-SA \cite{8897042}. {\color{black}For training each model, we use the code and the default settings for hyper-parameters, as mentioned in the respective papers. A short description of these papers is mentioned here for the sake of completeness and for a better understanding of the techniques experimented:
\begin{itemize}
    \item MVCNN - Uses two camera setups (12 views and 80 views) to render 2D images from a 3D model. These views are passed to a first CNN for extracting relevant features. The obtained features are then pooled and fed into a second CNN to obtain a compact shape descriptor.
    \item GVCNN - The 2D views of a 3D model are generated followed by a grouping of these views resulting in different clusters with associated weights. GoogLeNet is used as the base architecture.
    \item RotationNet - Uses only a partial set ($\geq$1) of the full multi-view images of an object as input. For each input image, the CNN also outputs the best viewpoint along with the predicted class, since the network treats the view-images as latent variables that are optimized in the training process.
    \item MVCNN-SA - Uses an approach similar to MVCNN, but attempts to assign relative importance to the input views by using an additional self-attention network.
\end{itemize}
} A train-test split ratio of 80\%-20\% is used on `CADSketchNet', which is similar to that of the MCB. The results of each of these methods are summarized in Section \ref{sec:res}.

\color{black}
\subsection{Experiments on Dataset-B}
\label{data-B}
Dataset-B in itself is not a large-scale dataset, and hence, not all deep network architectures can be trained on it. However, using the LFD images (1 3D CAD model = 20 images), the amount of available training data also increases. Hence, in addition to the view-based techniques mentioned above, we also come up with a few other rule-based and learning-based approaches and analyze their performance. The overall pipeline for performing a sketch-based search for 3D CAD models using deep learning can be broadly described as follows:
\begin{enumerate}
    \item Preparing a dataset of 3D CAD Models and their corresponding sketches suitable for training and testing a deep learning model (discussed in Section \ref{sec4}).
    \item Extracting feature representations from the CAD model as well as the query sketch.
    \item Developing the model architecture that can efficiently be trained using the extracted. representation(s) as input.
    \item Checking the similarity of the queried sketch and the 3D CAD models either directly or via their feature representations.
    \item Retrieving the top-ranked result(s) based on the metric(s) of similarity used.
\end{enumerate}
An overview of a generic sketch-based search engine for CAD models is shown in Fig. \ref{fig2}. In the following sub-sections, each step of the pipeline used is explained in greater detail.

\subsubsection{Feature Extraction and Model Architecture Details}
\label{sec5}
This section discusses steps 2 \& 3 of the pipeline, namely (1) extracting the feature representations of the query sketch and the 3D CAD models, and (2) using the extracted representations to build an appropriate network architecture. These two steps go hand-in-hand since the model architecture depends upon the dimensionality of the extracted features. If the extracted representation is a feature vector (1D), a deep neural network (DNN) or variational auto-encoders (VAE) can be used. If the extracted representations are images (2D), then a convolutional neural network (CNN), convolutional auto-encoder (CAE), or a Siamese Network (SN) are some possible options. In some other cases, the features are extracted by the neural network itself. The various methods experimented by us are described in this section.

\textbf{Model-1 : HOG-HOG}
Histogram of Oriented Gradients (HOG) is a widely used feature descriptor in computer vision and image processing. It differs from other feature extraction methods by extracting the edges' gradient and orientation rather than extracting the edges themselves. The input image is broken down into smaller localized regions, and for each region, the gradients and orientations are computed. Using these a Histogram is computed for each region. Our Model-1 uses the HOG in both Pipelines I and II (see Fig. \ref{fig2}).

\textit{Pipeline I:} The inputs to this pipeline are the sketch images from the dataset. These sketches are passed to the HOG algorithm, which generates the feature vectors for each image separately. The following configuration is used for the HOG algorithm after due experimentation: No. of pixels per cell: (8,8); No. of cells per block: (1,1); Orientations: 8; Block Normalization: L2; Feature Vector Size: 1024*1;

\textit{Pipeline II:} Using the idea of LFD, 20 images (256*256) are obtained for each 3D CAD model, resulting in 801*20 images. These images are then forwarded to the HOG block, which has a similar configuration as mentioned above, and the bag of features is obtained.

\textbf{Model-2 : HOG-AE}
\label{subsec2}
Model-2 uses the HOG pipeline as defined in Model-1 for Pipeline I and an auto-encoder for Pipeline II. The bag of features is obtained by training the auto-encoder (AE) on the LFD images and then extracting the encoded representation from the latent space of the AE. After various experiments, the following architecture for the AE is obtained. Encoder details: 8 conv layers with (3*3) filter and (1,1) stride; Batch Normalization (BN) is applied after every two conv layers; 2 dense layers; Decoder details: 2 dense and 8 deconv layers; an up-sampling layer of size (2,2) is applied after every two deconv layers; Activation function: LReLU with a negative slope of 0.01. This AE is trained for 30 epochs using the Adam Optimization algorithm. Learning rate: 0.0001; Loss: Mean Squared Error (MSE); 

\textbf{Model-3 : HOG-StackedAE}
\label{subsec3}
Model-3 uses the HOG pipeline as defined in Model-1 for Pipeline I and a stacked auto-encoder (SAE) for Pipeline II. Pipeline II is similar to the one defined in Model-2. Instead of passing all 801*20 images as separate inputs, the 20 images of each 3D CAD model are stacked and sent as a single input. Changes from pipeline in Model-2: Number of epochs: 50; Learning-rate: 3e-5;

\textbf{Model-4 : HOG-3DCNN}
\label{subsec3}
Model-4 uses the HOG pipeline as defined in Model-1 for Pipeline I and a 3D convolutional neural network (3D-CNN) for Pipeline II. For Pipeline II, each 3D CAD model is passed through a 3D-CNN. The extracted representations from the final dense layer are collected together as the bag of features. The architecture used for the 3D-CNN is 18 3D conv layers with kernel size (1,1,1) and stride (1,1,1); max-pooling with kernel-size=2, stride=2 along with BN applied after every 2 conv layers; average pool layer with kernel-size=2, stride=2; two dense layers with a dropout 0.5. This network is trained for 120 epochs with a learning rate of 0.0001; LReLU activation; Loss: MSE; Optimizer: Adam;

\renewcommand{\arraystretch}{1.1}
\begin{table*}[t]
\centering
\resizebox{0.9\textwidth}{!}{%
\begin{tabular}{|c|c|c|c|c|c|c|}
\hline
\textbf{Methodology} & \textbf{Training Time} & \textbf{Precision} & \textbf{Recall} & \textbf{Retrieval Time} & \textit{\textbf{mAP}} & \textbf{Top k-Accuracy \%} \\ \hline
MVCNN & 7h 30m  & 0.932 & 0.889 & 3.73$e$-05 & 0.894 & 94.03 \\ \hline
GVCNN & 7h 32m  & 0.959 & 0.904 & 3.79$e$-05 & 0.853 & 90.11 \\ \hline
RotationNet & 10h 14m & 0.947 & 0.868 & 3.78$e$-05 & 0.872 & 92.18 \\ \hline
MVCNN-SA & 7h 04m   & 0.941 & 0.903 & 3.69$e$-05 & 0.912 & 94.56 \\ \hline
\end{tabular}%
}
\caption{Results obtained by using various view-based approaches when trained using CADSketchNet - Dataset-A.}
\label{tab:expA}
\end{table*}
\renewcommand{\arraystretch}{1.00}

\renewcommand{\arraystretch}{1.1}
\begin{table*}[t]
\centering
\resizebox{.9\textwidth}{!}{%
\begin{tabular}{|c|c|c|c|c|c|c|}
\hline
\textbf{Methodology} & \textbf{Training Time} & \textbf{Precision} & \textbf{Recall} & \textbf{Retrieval Time} & \textit{\textbf{mAP}} & \textbf{Top k-Accuracy \%} \\ \hline
MVCNN  & 5h 22m	& 0.949 & 0.902 & 3.78$e$-05 & 0.912 & 95.15 \\ \hline
GVCNN  & 5h 25m & 0.972 & 0.894 & 3.85$e$-05 & 0.886 & 92.90 \\ \hline
RotationNet  & 8h 3m & 0.968 & 0.871 & 3.97$e$-05 & 0.909 & 96.66 \\ \hline
MVCNN-SA & 4h 58m & 0.9506 &	0.943 &	3.63$e$-05 & 0.947 & 96.23 \\ \hline
\end{tabular}%
}
\caption{Results obtained by using various view-based approaches when trained using CADSketchNet - Dataset-B.}
\label{tab:expB}
\end{table*}
\renewcommand{\arraystretch}{1.00}

\renewcommand{\arraystretch}{1.1}
\begin{table*}[t]
\centering
\resizebox{\textwidth}{!}{%
\begin{tabular}{|c|c|c|c|c|c|c|}
\hline
\textbf{Methodology} & \textbf{Training Time} & \textbf{Precision} & \textbf{Recall} & \textbf{Retrieval Time} & \textit{\textbf{mAP}} & \textbf{Top k-Accuracy \%} \\ \hline
HOG-HOG & -- & 0.666 & 0.031 & 3.72$e$-05 & 0.383 & 40.00\\ \hline
HOG-AE & 8h 20m  & 0.666 & 0.054 & 1.81$e$-05 & 0.526 & 51.25\\ \hline
HOG-StackedAE & 4h 30m & 0.666 & 0.020 & 2.04$e$-05 & 0.490 & 37.13\\ \hline
HOG-3DCNN & 10h 55m & 0.666 & 0.048 & 5.41$e$-05 & 0.458 & 41.21\\ \hline
Siamese Network (CNN-CNN) & 1h 10m & 0.970 & 0.784 & 2.88$e$-05 & 0.977 & 95.11\\ \hline
\end{tabular}%
}
\caption{Results obtained by using various models when trained on CADSketchNet - Dataset-B. The Siamese Network architecture outperforms other methods.}
\label{tab1}
\end{table*}
\renewcommand{\arraystretch}{1.00}

\textbf{Model-5 : CNN-CNN / Siamese Network}
\label{subsec3}
Model-5 uses a convolutional neural network (CNN) for both pipelines. This architecture is also known as Siamese Network, which implements the same network architecture and weights for both pipelines. The network is trained for 10 epochs; Learning Rate: 0.0001; Optimizer: Adam; Batch Size: 2; Loss: SiameseLoss function as described in \cite{Siamese}; Activation function: LeakyReLU with a negative slope = 0.01. 

To ensure that a sufficient proportion of similar and dissimilar pairs are generated, an approach similar to that of \cite{Siamese} is used. For each training sketch, a random number of view pairs ($k_p$) in the same category and $k_n$ view samples from other categories (dis-similar pairs) are chosen. In the current experiment, the values $k_p$ = 2, $k_n$ = 20 are used. This random pairing is done for each training epoch. For increasing the number of training samples, data augmentation for the sketch set is also done.

\section{Implementation Details}
\label{sec6}
\subsection{Coding Framework and System Configuration}
For implementing our neural network models, we use Python3 with PyTorch, while Python3 and sklearn were used to implement the HOG algorithm. OpenCV library is used for all implementations. All the implementations are carried out on a system running Ubuntu 18.04 Operating System. The system has an Intel Core i7-8700K CPU with 64GB RAM and an NVIDIA RTX 2080Ti GPU with 12GB RAM.

\subsection{Hyper-parameter Tuning, Loss function \& Optimization}
Training a neural network is an arduous process because of the many decisions involved beforehand, such as choice of performance metrics, hyper-parameters, loss function, etc. Our choices are mainly based on heuristics (\cite{Goodfellow, Bengio2012PracticalRF, Larochelle2009ExploringSF}) and are backed by experimental verification. The Weights \& Biases (wandb) library is also used to assist in hyper-parameter tuning.

\begin{figure*}
	\centering
	\includegraphics[scale=0.3]{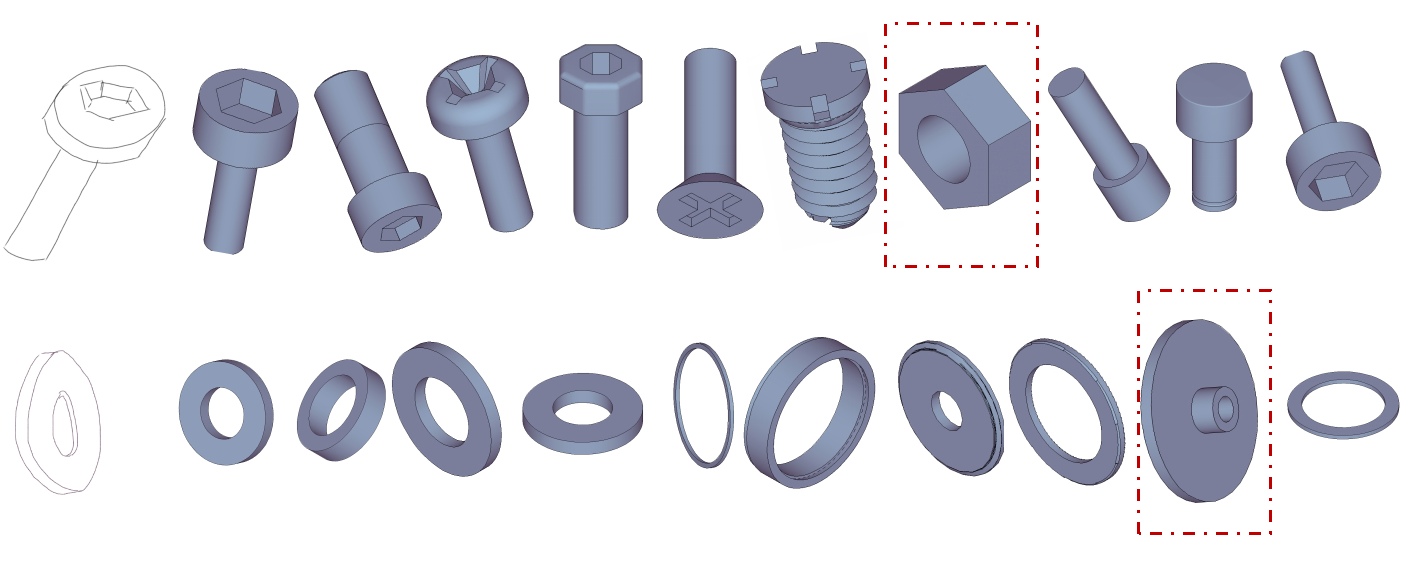}
	\caption{Top 10 search results of the developed Siamese network model for some sample input queries. Models retrieved in red boxes are from a different category.}
	\label{fig5}
\end{figure*}

\color{black}
\section{Results and Discussion}
\label{sec:res}
The results of all the Model experiments (see Section \ref{sec:exp}) are discussed here. Since there are no existing methods on the CADSketchNet dataset, we use these experimental results to compare the performance of the best retrieval system.  The results of search or retrieval are subjective and cannot be precisely quantified. Nevertheless, to evaluate the retrieval system's performance, we use the standard metrics popularly used in literature. The `top $k$ accuracy' denotes how many of the $k$-retrieved classes match the ground truth class. For instance, if 6 out of the top 20 retrieved results match the ground truth class, the accuracy is 30\%. For all the reported results, we use $k$=10. We also calculate the precision and recall values for the retrieval results. The mean Average Precision (mAP), which is the area under the P-R curve, is also computed. 

\renewcommand{\arraystretch}{1.02}
\begin{table*}[t]
\centering
\resizebox{0.8\textwidth}{!}{%
\begin{tabular}{|c|c|c|c|c|c|}
\hline
\textbf{CNN Architecture} & \textbf{Precision} & \textbf{Recall} & \textbf{Retrieval Time} & \textit{\textbf{mAP}} & \textbf{Top k-Accuracy \%} \\ \hline
LeNet5 \cite{LeNet} & 0.9222 & 0.6854 & 2.00$e$-05 & 0.6667 & 60.93\\ \hline
AlexNet \cite{AlexNet} & 0.8757 & 0.7746 & 2.22$e$-05 & 0.7901 & 72.11\\ \hline
VGG16 \cite{VGG} & 0.9112 & 0.6990 & 2.24$e$-05 & 0.6860 & 78.81\\ \hline
VGG16\_BN \cite{VGG} & 0.8989 & 0.7921 & 2.12$e$-05 & 0.7498 & 74.32\\ \hline
VGG19 \cite{VGG} & 0.8889 & 0.8100 & 2.20$e$-05 & 0.9004 & 88.80\\ \hline
VGG19\_BN \cite{VGG} & 0.9498 & 0.9100 & 2.60$e$-05 & 0.8905 & 83.42\\ \hline
Inceptionv3 \cite{Inceptionv3} & 0.9668 & 0.9310 & 2.90$e$-05 & 0.9457 & 90.31\\ \hline
DenseNet121 \cite{DenseNet} & 0.9579 & 0.9330 & 2.60$e$-05 & 0.9001 & 89.65\\ \hline
DenseNet161 \cite{DenseNet} & 0.9660 & 0.9320 & 2.85$e$-05 & 0.9220 & 91.90\\ \hline
DenseNet169 \cite{DenseNet} & 0.9410 & 0.9430 & 2.54$e$-05 & 0.9671 & 96.54\\ \hline
DenseNet201 \cite{DenseNet} & 0.9312 & 0.9256 & 2.78$e$-05 & 0.9256 & 95.73\\ \hline
Xception \cite{Xception} & 0.9790 & 0.9555 & 2.44$e$-05 & 0.9780 & 97.20\\ \hline
ResNet18 \cite{ResNet} & 0.9849 & 0.9610 & 3.56$e$-05 & 0.9660 & 97.50\\ \hline
ResNet34 \cite{ResNet} & 0.9516 & 0.9810 & 2.09$e$-05 & 0.9851 & 94.98\\ \hline
ResNet50 \cite{ResNet} & 0.9257 & 0.9066 & 2.67$e$-05 & 0.9234 & 92.30\\ \hline
ResNet101 \cite{ResNet} & 0.9445 & 0.9255 & 2.93$e$-05 & 0.9222 & 90.34\\ \hline
Resnet152 \cite{ResNet} & 0.9753 & 0.8923 & 3.34$e$-05 & 0.8231 & 96.80\\ \hline
ResNeXt50 \cite{ResNeXt} & 0.9407 & 0.9667 & 2.89$e$-05 & 0.9775 & 92.09\\ \hline
ResNeXt101 \cite{ResNeXt} & 0.9041 & 0.9433 & 2.85$e$-05 & 0.9432 & 88.91\\ \hline
\end{tabular}%
}
\caption{Results obtained by using various CNN architectures for the sketch-based retrieval of 3D CAD models when trained on the hand-drawn sketches of CADSketchNet (Dataset-B).}
\label{tab3}
\end{table*}
\renewcommand{\arraystretch}{1.00}

\color{black}
\subsection{Results of Experiments on Dataset-A}
The results of the four view-based methods experimented in Section \ref{data-A} are summarized in Table \ref{tab:expA} along with the time taken for retrieval. Both the MVCNN and the MVCNN-SA methods yield similar performance when tested on Dataset-A, with MVCNN-SA performing slightly better in terms of lesser training time, lesser time taken for retrieval and the top-$k$ accuracy. Although RotationNet takes the longest time for training, it performs better than GVCNN. This could be because of the following reasons. (1) GVCNN assigns relative weights for the input view images and groups the views according to the assigned weights. It might be that the viewing direction of the sketch image does not match with that of the group with the highest weight, thus slightly affecting the performance. (2) RotationNet takes into account the alignment of the input objects. Since object orientations in the MCB Dataset, and thereby the CADSketchNet-A, are aligned, the performance of the RotationNet model is enhanced.

\color{black}
\subsection{Results of Experiments on Dataset-B}
The results of the four view-based methods experimented in Section \ref{data-B} are summarized in Table \ref{tab:expB}. Also, for the experimental models described in \ref{sec5}, the outputs from pipelines I and II are compared using a `similarity check' block. For all the models (except Model-5), the Mean Squared Error (MSE) function is used to calculate the similarity between the extracted features. For Model-5, a custom Siamese Loss function described in \cite{Siamese} is used to measure the similarity. These results and the time taken for retrieval are reported in Table \ref{tab:expB}. 

\subsubsection{Discussion}
\label{subsec:disc}
Model-1 uses a simple HOG-HOG pipeline and is unable to provide the user with relevant search results. Other models use deep learning techniques and provide better results than the naive approach used in Model-1 (except Model-3). Even the time taken for retrieval is relatively higher, considering the inexpensive computational nature of an algorithmic approach instead of a data-driven approach.

Model-2 and Model-3 use a similar approach to retrieval, with the differences being Model-2 using the view images of the CAD models separately, while Model-3 uses the view images together (stacked). The time taken to train Model-3 (see Table \ref{tab1}) is expectedly lesser since Model-2 has to process 20 times more image data. While the results from Model-2 appear quite satisfactory (in the sense that the top-$k$ accuracy is slightly greater than 50\%), Model-3 severely under-performs. This might be because the network uses a stacked set of view images, and in an attempt to capture much information, the network overfits the data and thus under-performs. Thus Model-2, which treats each view image separately, performs much better.

Model-4 uses a 3DCNN, which is computationally very intensive. This is evident from Table \ref{tab1}, where Model-4 takes the highest time for training and searching. This is not conducive since, in real-time scenarios, the search results need to be delivered to the user quickly. Also, the model's performance is not satisfactory since capturing the 3D CAD models directly leads to unnecessary computations since most voxels in the input data would be empty (3D CAD models are typically sparse). This adversely affects the retrieval performance.

\color{black}

As it is evident from Table \ref{tab1}, using a Siamese network architecture (CNN in both pipelines) for the search engine yields the best results \textcolor{black}{among all methods that have been experimented, for all evaluation metrics.} Some sample search results are shown in Figure \ref{fig5}. Also, the model takes much lesser training time compared with other models. The search time depends a lot on the network architecture, the number of parameters to be trained and so forth. Also, the other methods use MSE for similarity measurement, while Model-5 uses a custom Siamese loss function which takes longer to compute. Moreover, search time is only a secondary measure of evaluating model performance. It is more important to obtain a good performance as opposed to obtaining inaccurate results quickly. 

Since Model-5 gives the best performance, various deep learning architecture pipelines are used for the Siamese network and are analyzed for performance. These results are reported in Table \ref{tab3}. It can be seen that using a ResNet18 backbone yields the best retrieval accuracy, and ResNet34 backbone results in the best $m$AP value. 

{\color{black} \textcolor{black}{In addition, the class-wise results of only the Model-5 are reported in Table \ref{tab:mcb_class} \& \ref{tab:mcb_class2} when trained on Dataset-A.} It is observed that the Siamese model performs very well on the computer-generated sketch Dataset-A. The top $k$-accuracy value for every class in the MCB dataset is in the range 97.06\% to 98.50\%. MCB is a large dataset and has sufficient examples in each category. Hence, the higher scores are justified.

Table \ref{tab:esb_class} reports the class-wise results of the Siamese Model when trained on Dataset-B, i.e. the hand-drawn sketches, \textcolor{black}{which has only 801 training samples. The results obtained differ significantly from those found for Dataset-A. The class 'Long Machine Elements,' which comprises only 15 sketches, has the lowest obtained accuracy rating of 86.67 percent. Considering the scarcity of training data,} this value indicates a good performance. Some categories only have single-digit training examples, such as `BackDoors' and `Clips'. Some of these classes obtain a 100\% accuracy, but this is only because there are insufficient number of testing examples. For a majority of the other categories, where there are a significant number of examples available to train, the Siamese Model performs quite well.}

\renewcommand{\arraystretch}{1.02}
\begin{table}[]
\centering
\resizebox{0.48\textwidth}{!}{%
\begin{tabular}{|c|c|c|c|}
\hline
\textbf{Method} & {\textbf{ModelNet40}} & {\textbf{Dataset-A}} & \textbf{Dataset-B}\\ \hline 
MVCNN       & 75.83\% & 94.03\% & 95.15\% \\ \hline 
GVCNN       & 82.46\% & 90.11\% & 92.90\% \\ \hline  
RotationNet & 86.84\% & 92.18\% & 96.66\% \\ \hline  
MVCNN-SA    & 85.94\% & 94.56\% & 96.23\% \\ \hline  
\end{tabular}%
}
\caption{The top k-accuracy values of deep-learning architectures that are pre-trained on ModelNet and tested on CADSketchNet vs. when trained on the developed CADSketchNet}
\label{tab:cg_cad}
\end{table}
\renewcommand{\arraystretch}{1.0}

\subsection{Comparison with deep-learning approaches used for 3D graphical models}
{\color{black} The results mentioned in the preceding sections are for models trained on the created CADSketchNet dataset, which focuses on 3D CAD models of mechanical components. The Section 1, Introduction, includes a description of how these CAD models differ from ordinary 3D shapes. In this section, we attempt to validate the assertion by employing deep-learning models designed for 3D graphical models. We examine the performance of the same view-based strategies that were mentioned in Section \ref{data-A}.} 

We use the same pipeline summarized in Figure \ref{fig2} for this purpose. \textcolor{black}{Pipeline I receives no changes, whereas Pipeline II employs the architecture that was pre-trained on ModelNet40 (a 3D graphical models dataset). The performance of the aforementioned approaches is then evaluated on the created CADSketchNet.} Table \ref{tab:cg_cad} summarizes these results. \textcolor{black}{Comparing these, with the methods trained on 3D CAD model data (Tables \ref{tab:expA} \& \ref{tab:expB}), it can be inferred that, the methodologies established for 3D graphical models do not translate well to 3D CAD models of engineering shapes. Network architectures trained on a dataset specific to CAD models perform significantly better. Therefore, instead of attempting to generalise the usage of approaches intended at graphical models upon engineering shapes, there is a need for developing dedicated datasets and methodologies that explicitly focus on 3D CAD models.}

\renewcommand{\arraystretch}{1.5}
\begin{table*}[]
\centering
\resizebox{\textwidth}{!}{%
\begin{tabular}{|c|c|c|c|c|c|c|c|}
\hline
\textbf{S No} & \textbf{Class} & \textbf{No.of Models} & \textbf{Precision} & \textbf{Recall} & \textbf{Retrieval Time} & \textbf{mAP} & \textbf{Top k-Accuracy} \\ \hline
1 &
  Articulations, eyelets and other articulated joints &
  1632 &
  0.992 &
  0.815 &
  4.10E-05 &
  0.986 &
  97.58 \\ \hline
2  & Bearing accessories          & 107  & 0.990 & 0.803 & 5.60$e$-05 & 0.985 & 98.11 \\ \hline
3  & Bushes                       & 764  & 0.988 & 0.820 & 4.20$e$-05 & 0.985 & 97.99 \\ \hline
4  & Cap nuts                     & 225  & 0.983 & 0.809 & 4.10$e$-05 & 0.989 & 97.84 \\ \hline
5  & Castle nuts                  & 226  & 0.979 & 0.820 & 4.80$e$-05 & 0.988 & 97.82 \\ \hline
6  & Castor                       & 99   & 0.983 & 0.821 & 4.60$e$-05 & 0.987 & 98.18 \\ \hline
7  & Chain drives                 & 100  & 0.974 & 0.814 & 4.00$e$-05 & 0.988 & 98.17 \\ \hline
8  & Clamps                       & 155  & 0.979 & 0.807 & 4.60$e$-05 & 0.989 & 98.21 \\ \hline
9  & Collars                      & 52   & 0.986 & 0.835 & 5.10$e$-05 & 0.984 & 97.56 \\ \hline
10 & Conventional rivets          & 3806 & 0.983 & 0.819 & 3.70$e$-05 & 0.986 & 97.82 \\ \hline
11 & Convex washer                & 91   & 0.981 & 0.818 & 3.40$e$-05 & 0.987 & 98.03 \\ \hline
12 & Cylindrical pins             & 1895 & 0.982 & 0.814 & 3.30$e$-05 & 0.985 & 97.69 \\ \hline
13 & Elbow fitting                & 383  & 0.992 & 0.823 & 3.10$e$-05 & 0.987 & 97.41 \\ \hline
14 & Eye screws                   & 1131 & 0.980 & 0.831 & 5.20$e$-05 & 0.984 & 98.16 \\ \hline
15 & Fan                          & 213  & 0.995 & 0.817 & 3.00$e$-05 & 0.989 & 97.15 \\ \hline
16 & Flanged block bearing        & 404  & 0.988 & 0.815 & 2.90$e$-05 & 0.988 & 98.03 \\ \hline
17 & Flanged plain bearings       & 110  & 0.985 & 0.816 & 4.30$e$-05 & 0.988 & 98.50 \\ \hline
18 & Flange nut                   & 53   & 0.991 & 0.808 & 5.60$e$-05 & 0.988 & 97.81 \\ \hline
19 & Grooved pins                 & 2245 & 0.980 & 0.812 & 3.50$e$-05 & 0.987 & 97.85 \\ \hline
20 & Helical geared motors        & 732  & 0.989 & 0.825 & 4.70$e$-05 & 0.986 & 97.83 \\ \hline
21 & Hexagonal nuts               & 1039 & 0.991 & 0.815 & 3.20$e$-05 & 0.988 & 97.50 \\ \hline
22 & Hinge                        & 54   & 0.976 & 0.815 & 5.60$e$-05 & 0.986 & 97.96 \\ \hline
23 & Hook                         & 119  & 0.985 & 0.828 & 5.40$e$-05 & 0.990 & 98.17 \\ \hline
24 & Impeller                     & 145  & 0.997 & 0.840 & 4.10$e$-05 & 0.989 & 97.06 \\ \hline
25 & Keys and keyways, splines    & 4936 & 0.993 & 0.818 & 6.10$e$-05 & 0.985 & 97.71 \\ \hline
26 & Knob                         & 644  & 0.988 & 0.809 & 4.20$e$-05 & 0.984 & 97.72 \\ \hline
27 & Lever                        & 1032 & 0.972 & 0.816 & 3.90$e$-05 & 0.987 & 97.66 \\ \hline
28 & Locating pins                & 55   & 0.992 & 0.820 & 4.70$e$-05 & 0.989 & 98.14 \\ \hline
29 & Locknuts                     & 254  & 0.988 & 0.805 & 5.30$e$-05 & 0.991 & 97.75 \\ \hline
30 & Lockwashers                  & 434  & 0.979 & 0.817 & 4.60$e$-05 & 0.986 & 98.04 \\ \hline
31 & Nozzle                       & 154  & 0.988 & 0.820 & 4.40$e$-05 & 0.988 & 97.77 \\ \hline
32 & Plain guidings               & 49   & 0.980 & 0.811 & 4.70$e$-05 & 0.987 & 98.25 \\ \hline
33 & Plates, circulate plates     & 365  & 0.985 & 0.813 & 2.20$e$-05 & 0.985 & 97.67 \\ \hline
34 & Plugs                        & 169  & 0.983 & 0.815 & 4.60$e$-05 & 0.983 & 98.03 \\ \hline
35 & Pulleys                      & 121  & 0.976 & 0.830 & 5.10$e$-05 & 0.988 & 97.99 \\ \hline
36 & Radial contact ball bearings & 1199 & 0.981 & 0.824 & 2.30$e$-05 & 0.988 & 97.83 \\ \hline
\end{tabular}%
}
\caption{Results obtained by the Siamese Network Architecture (Model-5) for the sketch-based retrieval of 3D CAD models when trained on the computer-generated sketches of CADSketchNet (Dataset-A): Part 1 of 2}
\label{tab:mcb_class}
\end{table*}

\begin{table*}[]
\centering
\resizebox{\textwidth}{!}{%
\begin{tabular}{|c|c|c|c|c|c|c|c|}
\hline
\textbf{S No} & \textbf{Class} & \textbf{No.of Models} & \textbf{Precision} & \textbf{Recall} & \textbf{Retrieval Time} & \textbf{mAP} & \textbf{Top k-Accuracy} \\ \hline
37 & Right angular gearings       & 60   & 0.984 & 0.829 & 5.10$e$-05 & 0.988 & 97.64 \\ \hline
38 & Right spur gears             & 430  & 0.991 & 0.815 & 3.80$e$-05 & 0.986 & 97.70 \\ \hline
39 & Rivet nut                    & 51   & 0.991 & 0.831 & 4.70$e$-05 & 0.984 & 97.63 \\ \hline
40 & Roll pins                    & 1597 & 0.990 & 0.814 & 4.20$e$-05 & 0.988 & 98.16 \\ \hline
41 & Screws and bolts with countersunk head & 2452  & 0.990 & 0.836 & 5.20$e$-05 & 0.982 & 97.46 \\ \hline
42 & Screws and bolts with cylindrical head & 3656  & 0.976 & 0.818 & 3.70$e$-05 & 0.988 & 98.06 \\ \hline
43 & Screws and bolts with hexagonal head   & 7058  & 0.983 & 0.819 & 6.10$e$-05 & 0.989 & 97.71 \\ \hline
44 & Setscrew                               & 1334  & 0.986 & 0.839 & 3.80$e$-05 & 0.983 & 97.93 \\ \hline
45 & Slotted nuts                           & 78    & 0.984 & 0.811 & 5.80$e$-05 & 0.989 & 98.03 \\ \hline
46 & Snap rings                             & 609   & 0.978 & 0.839 & 4.10$e$-05 & 0.986 & 97.86 \\ \hline
47 & Socket                                 & 858   & 0.983 & 0.816 & 3.80$e$-05 & 0.992 & 97.78 \\ \hline
48 & Spacers                                & 113   & 0.990 & 0.827 & 3.60$e$-05 & 0.987 & 98.03 \\ \hline
49 & Split pins                             & 472   & 1.000 & 0.830 & 5.10$e$-05 & 0.985 & 97.53 \\ \hline
50 & Springs                                & 328   & 0.986 & 0.819 & 5.30$e$-05 & 0.991 & 97.75 \\ \hline
51 & Spring washers                         & 55    & 0.991 & 0.819 & 2.60$e$-05 & 0.984 & 97.52 \\ \hline
52 & Square                                 & 72    & 0.991 & 0.818 & 5.10$e$-05 & 0.987 & 97.68 \\ \hline
53 & Square nuts                            & 53    & 0.987 & 0.821 & 4.40$e$-05 & 0.991 & 97.95 \\ \hline
54 & Standard fitting                       & 764   & 0.990 & 0.815 & 4.80$e$-05 & 0.987 & 98.23 \\ \hline
55 & Studs                                  & 4089  & 0.983 & 0.822 & 5.90$e$-05 & 0.988 & 97.55 \\ \hline
56 & Switch                                 & 173   & 0.984 & 0.816 & 5.70$e$-05 & 0.987 & 97.51 \\ \hline
57 & Taper pins                             & 1795  & 0.981 & 0.821 & 2.90$e$-05 & 0.987 & 98.15 \\ \hline
58 & Tapping screws                         & 2182  & 0.978 & 0.815 & 4.70$e$-05 & 0.986 & 97.70 \\ \hline
59 & Threaded rods                          & 1022  & 0.979 & 0.811 & 3.10$e$-05 & 0.987 & 97.99 \\ \hline
60 & Thrust washers                         & 2333  & 0.992 & 0.835 & 3.20$e$-05 & 0.985 & 98.14 \\ \hline
61 & T-nut                                  & 101   & 0.980 & 0.823 & 4.40$e$-05 & 0.985 & 97.81 \\ \hline
62 & Toothed                                & 47    & 0.989 & 0.815 & 5.30$e$-05 & 0.988 & 97.85 \\ \hline
63 & T-shape fitting                        & 338   & 0.975 & 0.812 & 1.60$e$-05 & 0.987 & 98.03 \\ \hline
64 & Turbine                                & 85    & 0.979 & 0.823 & 5.30$e$-05 & 0.986 & 97.63 \\ \hline
65 & Valve                                  & 94    & 0.981 & 0.815 & 3.30$e$-05 & 0.989 & 97.67 \\ \hline
66 & Washer bolt                            & 912   & 0.982 & 0.837 & 4.80$e$-05 & 0.986 & 97.57 \\ \hline
67 & Wheel                                  & 243   & 0.979 & 0.826 & 4.20$e$-05 & 0.989 & 97.47 \\ \hline
68 & Wingnuts                               & 50    & 0.992 & 0.816 & 5.40$e$-05 & 0.989 & 97.81 \\ \hline
  & Overall                                & 58696 & 0.985 & 0.820 & 4.34$e$-05 & 0.987 & 97.83 \\ \hline
\end{tabular}%
}
\caption{Results obtained by the Siamese Network Architecture (Model-5) for the sketch-based retrieval of 3D CAD models when trained on the computer-generated sketches of CADSketchNet (Dataset-A): Part 2 of 2}
\label{tab:mcb_class2}
\end{table*}
\renewcommand{\arraystretch}{1.0}

\renewcommand{\arraystretch}{1.2}
\begin{table*}[]
\centering
\resizebox{\textwidth}{!}{%
\begin{tabular}{|c|c|c|c|c|c|c|c|}
\hline
\textbf{S No} & \textbf{Class} & \textbf{No.of Models} & \textbf{Precision} & \textbf{Recall} & \textbf{Retrieval Time} & \textbf{mAP} & \textbf{Top k-Accuracy} \\ \hline
1  & 90 degree elbows      & 41  & 0.964 & 0.853 & 2.21$e$-05 & 0.960 & 95.12  \\ \hline
2  & BackDoors               & 7   & 0.927 & 0.571 & 1.92$e$-05 & 0.957 & 100.00 \\ \hline
3  & Bearing Blocks         & 7   & 0.984 & 0.714 & 2.78$e$-05 & 0.967 & 100.00 \\ \hline
4  & Bearing Like Parts    & 20  & 0.991 & 0.900 & 2.83$e$-05 & 0.964 & 90.00  \\ \hline
5  & Bolt-like Parts       & 53  & 0.975 & 0.894 & 2.80$e$-05 & 0.966 & 92.45  \\ \hline
6  & Bracket-like Parts    & 18  & 0.972 & 0.778 & 2.68$e$-05 & 0.978 & 94.44  \\ \hline
7  & Clips                   & 4   & 0.987 & 0.500 & 2.18$e$-05 & 0.987 & 100.00 \\ \hline
8  & Contact Switches       & 8   & 0.971 & 0.750 & 2.47$e$-05 & 0.971 & 87.50  \\ \hline
9  & Container-like Parts  & 10  & 0.981 & 0.700 & 2.76$e$-05 & 0.992 & 100.00 \\ \hline
10 & Contoured Surfaces     & 5   & 0.962 & 0.600 & 2.33$e$-05 & 0.986 & 100.00 \\ \hline
11 & Curved Housings        & 9   & 0.972 & 0.778 & 2.18$e$-04 & 0.972 & 100.00 \\ \hline
12 & Cylindrical Parts      & 43  & 0.980 & 0.930 & 2.88$e$-05 & 0.974 & 93.02  \\ \hline
13 & Discs                   & 51  & 0.970 & 0.824 & 3.03$e$-05 & 0.977 & 94.12  \\ \hline
14 & Flange-like Parts     & 14  & 0.979 & 0.786 & 2.12$e$-05 & 0.978 & 92.86  \\ \hline
15 & Gear-like Parts       & 36  & 0.976 & 0.833 & 2.89$e$-05 & 0.985 & 97.22  \\ \hline
16 & Handles                 & 18  & 0.963 & 0.889 & 1.92$e$-05 & 0.974 & 100.00 \\ \hline
17 & Intersecting Pipes     & 9   & 0.962 & 0.667 & 2.78$e$-05 & 0.956 & 100.00 \\ \hline
18 & L-Blocks               & 7   & 0.951 & 0.714 & 2.38$e$-05 & 0.979 & 100.00 \\ \hline
19 & Long Machine Elements & 15  & 0.974 & 0.800 & 2.91$e$-05 & 0.978 & 86.67  \\ \hline
20 & Long Pins              & 58  & 0.969 & 0.927 & 2.63$e$-05 & 0.976 & 94.83  \\ \hline
21 & Machined Blocks        & 9   & 0.965 & 0.778 & 2.48$e$-05 & 0.966 & 100.00 \\ \hline
22 & Machined Plates        & 49  & 0.984 & 0.898 & 2.23$e$-05 & 0.986 & 91.84  \\ \hline
23 & Motor Bodies           & 19  & 0.983 & 0.842 & 1.97$e$-05 & 0.987 & 94.74  \\ \hline
24 & Non-90 degree elbows  & 8   & 0.973 & 0.875 & 2.61$e$-05 & 0.975 & 87.50  \\ \hline
25 & Nuts                    & 19  & 0.971 & 0.737 & 1.99$e$-05 & 0.972 & 95.11  \\ \hline
26 & Oil Pans               & 8   & 0.970 & 0.750 & 2.68$e$-05 & 0.963 & 89.47  \\ \hline
27 & Posts                   & 11  & 0.972 & 0.728 & 2.76$e$-05 & 0.984 & 90.91  \\ \hline
28 & Prismatic Stock        & 36  & 0.976 & 0.861 & 2.88$e$-05 & 0.977 & 94.44  \\ \hline
29 & Pulley-like Parts     & 12  & 0.974 & 0.833 & 2.98$e$-05 & 0.982 & 91.67  \\ \hline
30 & Rectangular Housings   & 7   & 0.968 & 0.571 & 2.77$e$-05 & 0.983 & 100.00 \\ \hline
31 & Rocker Arms            & 10  & 0.980 & 0.700 & 1.62$e$-05 & 0.991 & 100.00 \\ \hline
32 & Round Change At End  & 21  & 0.953 & 0.857 & 1.91$e$-05 & 0.967 & 95.24  \\ \hline
33 & Simple Pipes           & 16  & 0.956 & 0.875 & 2.21$e$-05 & 0.959 & 93.75  \\ \hline
34 & Slender Links          & 13  & 0.985 & 0.769 & 2.24$e$-05 & 0.987 & 100.00 \\ \hline
35 & Slender Thin Plates   & 12  & 0.971 & 0.750 & 2.13$e$-05 & 0.979 & 95.11  \\ \hline
36 & Small Machined Blocks & 12  & 0.962 & 0.833 & 2.42$e$-05 & 0.985 & 100.00 \\ \hline
37 & Spoked Wheels          & 15  & 0.959 & 0.867 & 2.65$e$-05 & 0.978 & 93.33  \\ \hline
38 & T-shaped parts         & 15  & 0.954 & 0.800 & 2.48$e$-05 & 0.986 & 86.67  \\ \hline
39 & Thick Plates           & 23  & 0.965 & 0.826 & 1.99$e$-05 & 0.971 & 91.30  \\ \hline
40 & Thick Slotted plates    & 15  & 0.969 & 0.733 & 1.71$e$-05 & 0.987 & 93.33  \\ \hline
41 & Thin Plates            & 12  & 0.971 & 0.833 & 2.19$e$-05 & 0.995 & 100.00 \\ \hline
42 & U-shaped parts         & 25  & 0.972 & 0.800 & 1.82$e$-05 & 0.992 & 92.00  \\ \hline
  & Overall & 801 & 0.970 & 0.784 & 2.88$e$-05 & 0.977 & 95.11  \\ \hline
\end{tabular}%
}
\caption{Results obtained by the Siamese Network Architecture (Model-5) for the sketch-based retrieval of 3D CAD models when trained on the hand-drawn sketches of CADSketchNet (Dataset-B).}
\label{tab:esb_class}
\end{table*}
\renewcommand{\arraystretch}{1.00}

\subsection{Limitations and Possible future work}
\textcolor{black}{The scope of this work is confined to 3D Engineering CAD Mesh models. The current method does not retrieve other types of data, such as images or 3D point sets. It is worthwhile to investigate the possibility of creating a unified dataset with multiple input formats and developing a retrieval system that can search for multiple data formats simultaneously.} Also, only sketch query inputs are handled by the proposed model and not text query or 3D model query. 

\begin{figure}
\includegraphics[width=0.48\textwidth,height=4.5cm]{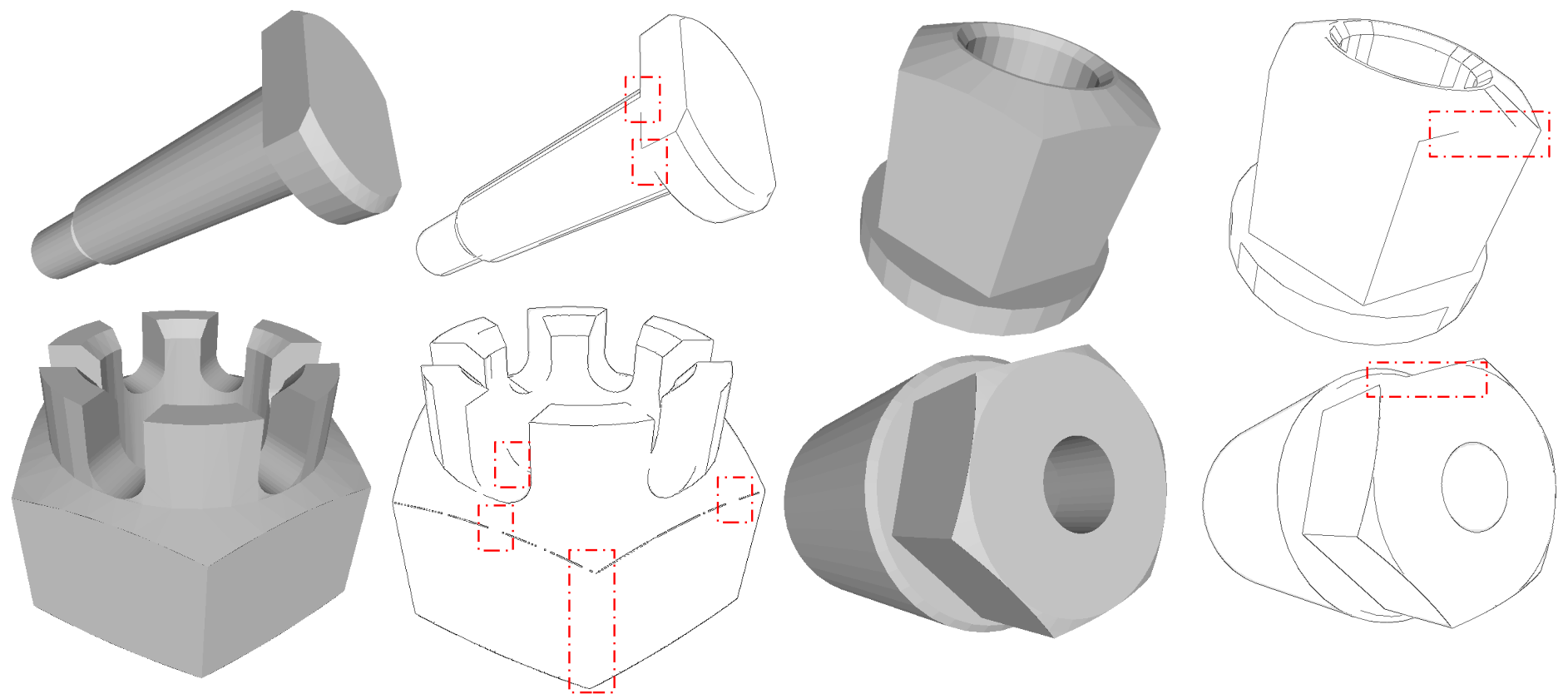}
\caption{Illustrating some sample cases of improperly generated computer-sketch data by the proposed method.}
\label{fig:fail}
\end{figure}

{\color{black} Some computer-generated sketches have unintended artifacts caused by image processing algorithms and a few sketches} miss out on some features  (Figure \ref{fig:fail}). While additional processing or other complex sketch generation \textcolor{black}{techniques might potentially be utilised for such sketches, it could considerably increase the time required to gather large-scale data.}

As far as the hand-drawn sketches are concerned, since the users are primarily from the design background, the obtained sketches are in the same context. {\color{black} In future, gathering sketches from the perspectives of other area experts, such as assembly or machining, could be considered. 

Another possible future work would be for people to contribute to the dataset by providing hand-drawn sketches, which may be used to increase the size of the dataset once it is made open.} Recent 3D CAD model datasets such as \cite{CADNET} can also be utilised to generate more sketch images, and the dataset \textcolor{black}{can be further increased.} The current work could also be extended to a CAD assembly model retrieval \cite{CAD_assem}.

\section{Conclusion}
\label{sec:concl}
A sketch dataset of computer-generated query images, called `CADSketchNet', has been built using the available 3D CAD models from the MCB dataset and the ESB dataset. These images, along with each 3D CAD model's representative images, are stored in a database. Additionally, hand-drawn sketches corresponding to CAD models from the ESB dataset are also created and included in `CADSketchNet'. This dataset will be made open-source and could contribute to the development of image-based search engines for 3D mechanical component CAD models - using the latest advances in deep learning. The performance of standard view-based methods proposed in the literature is analyzed on CADSketchNet. Additionally, the results of a few other experiments using popular deep learning architectures are also reported. The possibilities of extending this research work to other similar problems have also been discussed.

\section*{Acknowledgments}
Thanks are due to the teams of the ESB and the MCB datasets for making their data publicly available. Thanks are also due to many users who have contributed to our CADSketchNet dataset.

\footnotesize
\bibliographystyle{unsrt}
\bibliography{refs}
\end{document}